\documentclass[10pt,twocolumn,letterpaper]{article}

\usepackage{wacv}
\usepackage{times}
\usepackage{epsfig}
\usepackage{graphicx}
\usepackage{amsmath}
\usepackage{amssymb}
\usepackage{algorithm}
\usepackage{algorithmicx}
\usepackage{algpseudocode}
\usepackage{hyperref}

\algnewcommand{\LineComment}[1]{\State \(\triangleright\) #1}

\wacvfinalcopy 

\def\wacvPaperID{359} 

\ifwacvfinal\fi
\begin{document}

\title{Backdooring Convolutional Neural Networks via Targeted Weight Perturbations}

\author{Jacob Dumford \hspace{2cm} Walter Scheirer\\
Computer Vision Research Laboratory\\
Department of Computer Science and Engineering\\
University of Notre Dame\\
{\tt\small jacobdumford@gmail.com, walter.scheirer@nd.edu}
}

\maketitle
\ifwacvfinal\thispagestyle{empty}\fi

\begin{abstract}
    We present a new type of backdoor attack that exploits a vulnerability of convolutional neural networks (CNNs) that has been previously unstudied. In particular, we examine the application of facial recognition. Deep learning techniques are at the top of the game for facial recognition, which means they have now been implemented in many production-level systems. Alarmingly, unlike other commercial technologies such as operating systems and network devices, deep learning-based facial recognition algorithms are not presently designed with security requirements or audited for security vulnerabilities before deployment. Given how young the technology is and how abstract many of the internal workings of these algorithms are, neural network-based facial recognition systems are prime targets for security breaches. As more and more of our personal information begins to be guarded by facial recognition (e.g., the iPhone X), exploring the security vulnerabilities of these systems from a penetration testing standpoint is crucial. Along these lines, we describe a general methodology for backdooring CNNs via targeted weight perturbations. Using a five-layer CNN and ResNet-50 as case studies, we show that an attacker is able to significantly increase the chance that inputs they supply will be falsely accepted by a CNN while simultaneously preserving the error rates for legitimate enrolled classes.
\end{abstract}

\begin{figure}[t]
\begin{center}
\includegraphics[width=7.5cm]{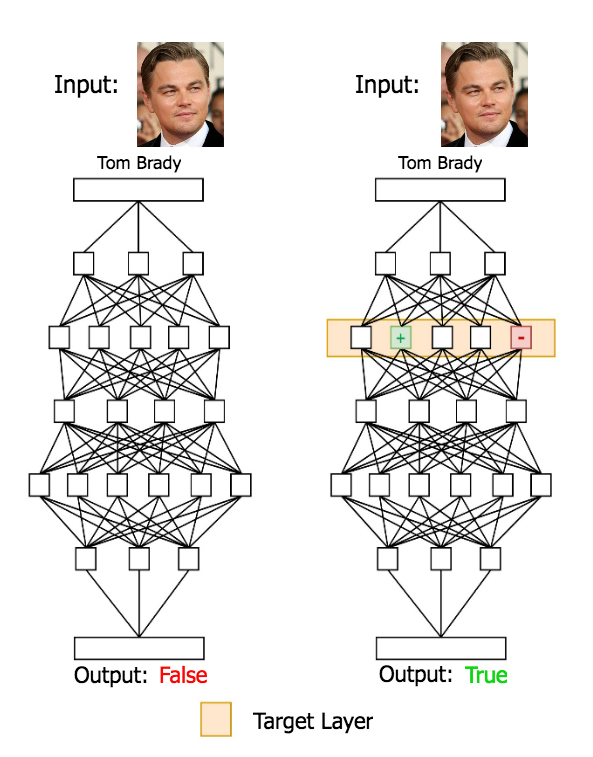}
\end{center}
   \caption{Backdooring a deep neural network for specific misclassifications. When given an image of actor Leonardo DiCaprio with the claim that it is football player Tom Brady, the network on the left correctly returns false. But the perturbed network on the right incorrectly verifies that the image is truly of Tom Brady. Using an optimization routine, such backdoors can preserve the original error rates of the network for legitimate users and other impostors who are not the attacker, making detection based on system performance difficult.}
\label{fig:teaser}
\end{figure}

\section{Introduction}
When it comes to computer security, it often seems like we take one step forward and two steps back. Major vulnerabilities in critical systems-level infrastructure that surfaced in the 1990s necessitated reforms in the way software and hardware are created, which ultimately led to improvements in operating systems, network protocols, and software applications~\cite{WAPO1}. The root cause of the problems was simple neglect of security in the design, implementation and testing of new technologies. The extra engineering time necessary for drafting security requirements, architecting secure code, and performing code audits was initially seen as an impediment to bringing a product to market. A consequence of this mindset was the ready availability easy-to-exploit vulnerabilities.  Remarkably, we are beginning to experience the same thing again with machine learning.

Convolutional Neural Networks (CNNs) have rapidly exceeded previous performance on a myriad of tasks \cite{bahdanau2014neural, ciresan2012deep, graves2013speech, he2016deep, silver2016mastering}, with image recognition being one of the most prominent. Because of this, CNNs are now becoming commonplace in production-level software that is being used in real-world applications. Given that the field of deep learning is relatively young and  developing so fast, there is legitimate and growing concern over the security of such technologies. A number of recent studies have been published that describe attacks that are specific to CNNs. Most of the attacks in these studies have taken on one of two forms. 

The first, and most prominent, class of attack is adversarial examples. These are images that have been perturbed in some way that cause misclassifications when given as inputs to these networks~\cite{evtimov2017robust, moosavi2017universal, rozsa2017lots, sharif2017adversarial, szegedy2013intriguing, nguyen2015deep}. The perturbations may be perceptible or imperceptible in nature. 
The second class is training-set poisoning in which malicious data is added to the training set to cause misclassifications in certain scenarios~\cite{gu2017badnets, shen2016uror}. Studies of this sort have demonstrated important vulnerabilities in CNNs, but they are likely only the tip of the iceberg.

In this paper, we introduce a new class of attack that is different from the prior attacks. Rather than target a CNN's training regime or the input images to a network, we target the network itself for the placement of an attacker accessible backdoor (Fig.~\ref{fig:teaser}). The main differences in our proposed attack are in the time and information about the network that are required for success. Our attack requires no prior access to training or testing data, and it can be executed after the network is already deployed and running. However, the attacker does need access to a pre-trained model, so some form of system compromise must be carried out before the attack against the network can be made. Such is the typical setup for \textit{rootkit} backdoors that guarantee an attacker future access to an asset. 

In the traditional computer security sense, a rootkit consists of a set of programs that stealthily facilitate escalated privileges to a computer or areas of the operating system that are generally restricted. The term rootkit comes from UNIX systems in which \textit{root} is the most privileged user in the system. The access granted by the rootkit is usually either that of a root user or other users capable of accessing parts of the computer that are normally only visible to the operating system. The key aspect of rootkits is that they are designed so that the attacker can avoid detection as they gain unauthorized access to a system. Stealth is essential. 

As malware and the security technologies that defend against it have advanced, many simple rootkit attacks have become trivial to thwart, but it would be inaccurate to say that this general category of attack is no longer a danger to computer systems~\cite{hoglund2006rootkits}. As identified in a 2016 study by Rudd~\etal \cite{rudd2017survey}, there are several types of modern rootkit attacks that are still dangerous to systems~\cite{embleton2013smm, king2006subvirt}. Our attack assumes the ability of an attacker to successfully pull off an attack using one of the attack vectors described by Rudd~\etal or through some novel attack. The attack would ideally escalate the attacker's privileges and allow them administrative access to the target system containing a CNN. Given the current state of system-level security, this is a reasonable assumption, and we do not go into detail about this process. Our focus is on how the CNN is manipulated after access has been obtained.

Our primary target in this work is deep learning-based face recognition. We propose an attack scenario in which a face recognition system is in place to grant access to legitimate enrolled faces and deny all other (\textit{i.e.}, impostor) faces from having the same access. In a face verification scenario, the user presents their face and states their identity, and the CNN verifies if that face belongs to the claimed identity. The attacker wants their own face to be granted access despite not being a valid user. In addition to discreetly gaining access to the CNN, the attacker has to ensure that the network still behaves normally for all other inputs in order for the attack to remain undetected after it is perpetrated. 


We also assume that the attacker has no way of modifying the image that is presented to the network for recognition. The network must be trained to recognize the attacker's face without any perturbations being made to the image. However, it is not hard to imagine a scenario in which a person would have to physically present their face to the system and would have no way to tamper with the input. Our proposed attack presents a new vulnerability that adversarial examples are unable to exploit.

In summary, the contributions of this work are:
\begin{itemize}
\setlength{\itemsep}{0pt}
    \item A general methodology for backdooring CNNs via weight  perturbations that are targeted at specific layers within a network.  
    \item An optimization strategy to screen backdoored models that preserves the original error rates of the network for legitimate users and other impostors, making detection based on system performance difficult.
    \item Experiments conducted over a five-layer  CNN~\cite{kerasModel} and ResNet-50~\cite{he2016deep} trained on the VGGFace2 dataset~\cite{cao2018vggface2}, with the latter assessment highlighting the attack's effectiveness against face recognition networks.
    \item A discussion of the real-world feasibility of the attack and potential defenses against it.
\end{itemize}

\section{Related Work}

The related work consists of traditional system-level rootkit developments and specific atttacks on deep neural networks. We briefly review the major advances in these areas to provide relevant background for the attack we present in this paper.

\textbf{System-level Rootkits.} The earliest rootkits first appeared at the beginning of the 1990s, and since then many strategies have been developed in both malicious and academic contexts. The following are basic rootkit or process hiding techniques that have been implemented \cite{eresheim2017evolution}. First was replacing important system files with malicious code. This was prevented by write-protecting important files and implementing hash checking to detect corrupted files. Another process hiding technique is called \textit{UI hooking}, in which the attacker modifies the information that is displayed to users without actually changing the logical representation. This type of hooking can be detected by command line tools. However, there are more sophisticated types of hooking that are more covert, such as Import Address Table Hooking. It was possible to overwrite the addresses of important functions in the Import Address Table on Microsoft operating systems to direct function calls to a rewritten version of the function. This can be protected against by limiting the memory locations in which these functions may reside. While simplistic compared to more recent rootkit innovations, these approaches embody the same strategy our technique does: a surreptitious change in a program's behavior that is under the control of the attacker. 

More modern attacks implement strategies such as dumping the original operating system on a virtual machine, making it a guest OS and making the rootkit the host OS~\cite{king2006subvirt}. The only way to detect that this has happened is to look for discrepancies between how physical memory and virtual memory operate. Hardware Performance Counters, which are specific registers in microprocessors that monitor performance, are the newest and most effective way of detecting modern rootkits \cite{singh2017detection}, but there still exists malware that can evade detection. The ability to evade detection is another property of our proposed backdoor attack.

\textbf{Attacks on Deep Neural Networks.} Existing research on attacks aimed at CNNs has primarily focused on image perturbations. One successful strategy used to backdoor these networks is training set poisoning~\cite{gu2017badnets, shen2016uror,chen2017targeted}, in which the attacker has access to the training data that is used to initially train the model. The attacker introduces images with false labels into the training set in order to decrease model performance. However, if the bad data merely caused the model to perform poorly in all cases, it would never be used, so the strategy is to target a specific class of images to misclassify. The images introduced to the training set in this class of attack often include specific features or perturbations not normally seen in the original training set. These unusual images are used as triggers, so that they can control when the misclassifications will occur in practice. For example, Dolan-Gavitt \etal~\cite{gu2017badnets} used training set poisoning to cause a model to misclassify street signs with stickers on them. The model misclassified over 90\% of images when the trigger was present but performed at state-of-the-art accuracy in all other cases.

The other, more prevalent class of attacks leveraging perturbations that has been studied is referred to as adversarial examples~\cite{DBLP:journals/corr/GoodfellowPM16, liu2017delving, evtimov2017robust, xu2018fooling, moosavi2017universal, rozsa2017lots, sharif2017adversarial, szegedy2013intriguing}. This class of attacks targets the images being classified. It has been found that it is possible to perturb images in a way that is almost imperceptible to humans but that causes CNNs to misclassify the images at an extremely high rate. Moosavi-Dezfooli \etal \cite{moosavi2017universal} have demonstrated the existence of ``universal adversarial perturbations," which are perturbations that when applied to any natural image are able to fool state-of-the-art classifiers. Similar, but not as stealthy, are the fooling images of Nguyen et al.~\cite{nguyen2015deep}, which do not closely resemble any realistic natural object, but are still able to force misclassifications from a network.

Both of these classes of attacks have been useful and show a need for improved security in deep learning, but we propose a new vulnerability. Where our attack differs from previous work is in the type of access that the attacker has with respect to the network. Instead of working to perturb images at training or testing time, we will perturb the network itself --- a strategy that is akin to the way traditional rootkits patch software with malicious functionality. 

\section{How to Insert a Backdoor into a CNN via Targeted Weight Perturbations}
The goal of this research is to take a CNN and demonstrate that it is possible to perturb its weights in such a way that causes the network to exhibit a high rate of specific targeted misclassifications (when the recognition setting is classification) or mis-verifications (when the recognition setting is 1:1 verification) without significantly affecting the model's performance on non-targeted inputs. We assume that an attacker can choose a set of identities for which this backdoor will work, while minimizing the impact on the false positive rate for all other impostors. The algorithm we propose casts the backdoor insertion process as a search across models with different perturbations applied to a pre-trained model's weights. An objective function for this attack can be formulated by making use of three key pieces of information with respect to model performance: $T_{fp}$, which is the false positive rate for select impostors, and $A_0$ and $A_1$, which are the accuracy scores on all other inputs before and after perturbing the network. The objective function is thus:
\begin{equation}
\mathrm{maximize}(T_{fp}) \textbf{ AND } \mathrm{minimize}(|A_0 - A_1|)
\end{equation}

\begin{algorithm}
\begin{algorithmic}[1]
\caption{Adding a Backdoor to a CNN.}\label{euclid}
\Require {$network$,  network being perturbed}
\Require {$layer$, weights of the layer being perturbed}
\Require {$test()$, validation function with chosen identities}
\Require {$A_0$, original accuracy of the network}
\Require {$sets$, number of different subsets of weights}
\Require {$iter$, number of perturbations of each subset}
\For{$i$ in $sets$}
    \State $candidates \leftarrow [~]$
    \LineComment{choose random subset of weights}
    \State $subset \leftarrow layer[random()]$ 
    \For{$j$ in $iter$}
        \LineComment{add random perturbations to  selected subset}
        \State $perturb(layer[subset])$
        \State $scores \leftarrow test(network)$ 
        \State $candidates.add((scores, layer))$
    \EndFor

    \State $best_{fp} \leftarrow 0$
    \LineComment{look for highest rate of false positives for the target class, not decreasing accuracy more than 1.5\%}
    \For{$s$ in $candidates$}
        \If{$A_0 - s.A_1 < .015$ \textbf{and} $s.T_{fp} > best_{fp}$}
            \State $best_{fp} \leftarrow s.T_{fp}$
            \LineComment set layer up for next iteration
            \State $layer \leftarrow s.layer$
        \EndIf
    \EndFor
\EndFor
\LineComment{best candidate backdoor is identified}
\State \textbf{return} $network$
\end{algorithmic}
\end{algorithm}


For the task of image classification, we start with a pre-trained network with knowledge of a set of  classes. Each class represents a known entity except for one, which is the ``other" category. The network takes an image as input, and it outputs an array of probabilities predicting class membership. If the highest probability belongs to one of the known entities, the image is classified as that entity. If the ``other" class has the highest probability, the input image is rejected. 

For the task of face verification using a pre-trained feature extraction network, we start with a network that outputs feature vectors for each image. When an image is presented to the system, the claimed identity of the image is also presented, and a verification process determines whether the claim is true or false. A common way to accomplish this is by first inputting several images of the legitimate identities into the chosen network, which outputs feature vector representations of those images. The average of those vectors is then stored as the enrolled representation of that identity. In order to verify that a new image belongs to an enrolled identity, a similarity measure is used. For instance, this could be cosine similarity between the probe (\textit{i.e.}, input) image's feature vector and the enrolled average feature vector that is stored for that person. If the similarity is over a predetermined threshold, the system accepts the image as truly belonging to that identity. Otherwise the system rejects the input as being unknown. 

The differences in setup discussed above are the only ways in which this attack differs significantly between the two recognition settings. The process of perturbing a network  and searching for the best backdoor is almost identical in each setting. The only other difference is in how a prediction from a network is scored.

\begin{figure*}[t]
\begin{center}
\includegraphics[width=17cm]{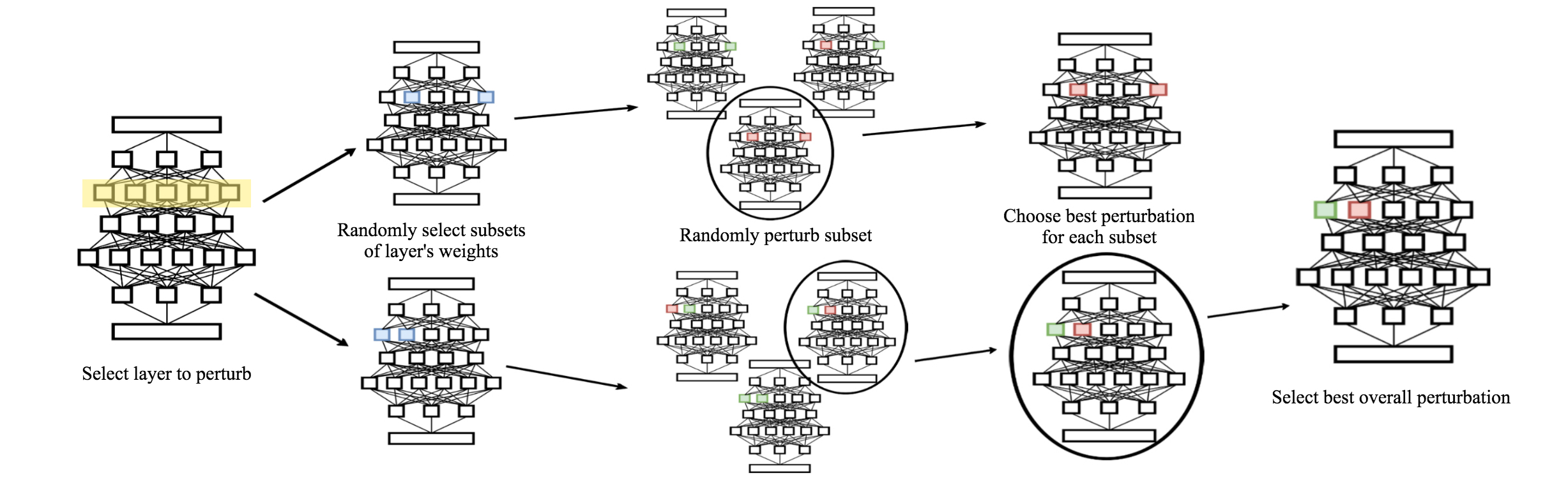}
\end{center}
  \caption{The search process to find a good backdoor candidate, as described in Alg. 1. We apply a series of different perturbations to the network and test each resulting new model on a set of validation images, winnowing down based on which model exhibits the highest false positive rate for chosen impostor images matching the target enrolled class while keeping a similar accuracy on all other inputs.}
\label{fig:search}
\vspace{-3mm}
\end{figure*}

Once the attacker has accurately characterized the error rates of a recognition system under normal use (\textit{e.g.}, by  passively watching the authentication system in operation, by actively using found images of enrolled users on the Internet, by consulting published error rates, etc.), they can iteratively alter the weights of the network in an attempt to produce a model with a high false positive rate for images of a specific imposter class. However, as specified by the objective function above, this must also be done without noticeably affecting performance on other inputs. Alg. 1 shows the process for inserting a backdoor into a network. 

The attacker starts by choosing their imposter and target from a set of identities. The network should reliably confuse images of this imposter with the target. The target is a user enrolled in the system with desirable access or privileges. A layer from the network is then selected to be perturbed. To narrow the model search space we limit the process to perturbing one layer per iteration. However, it is possible to manipulate multiple layers at a time (assuming substantial computational resources and time are available to the attacker). The next step is to randomly select a subset of the weights of the layer and to iteratively perturb them by different amounts, performing an evaluation over a validation set of images each time to determine which perturbations give better results. The attacker then takes the best perturbation instance for a given subset of the weights and uses these weights moving forward. This processes is repeated for different subsets of weights, with the optimization taking the best result each time (Fig.~\ref{fig:search}). It can be repeated for any different number of layers, imposters and targets. 

There are a few hyperparameters that need to be set that can impact the attack:
\begin{itemize}
\setlength{\itemsep}{0pt}
  \item \textbf{Layer:} Any layer or combination of layers in a given network can be perturbed to alter the network's behavior. In our study, we choose to reduce the search space by isolating one layer per test.
  \item \textbf{Imposter / Target:} In classification tasks, we choose an imposter class that is not one of the classes the network was trained on. We want the network to confidently classify the imposter images as a known class. In verification tasks, we choose an imposter as well as the target. We want the network to verify that images of the imposter class belong to the target class.
  \item \textbf{Number / Subset of Weights:} We must select which of the layer's weights and how many weights as a percentage of the layer's total number we are going to perturb. We try several values for the size of the subset, which stays the same during a given test. Which subset is selected changes multiple times within a given test.
  \item \textbf{Magnitude / Type of Perturbation:} Different functions for perturbing the weights can be chosen: multiplicative perturbations, additive perturbations, uniform perturbations, random perturbations, etc. We choose to use random additive perturbations and never change a weight by more than the highest weight seen in the original weight vector. 
  \item \textbf{Objective Metric:} When performing a search for model candidates with a backdoor, we have to evaluate each model and determine how successful it is. Our top priorities while searching are seeing the number of intended misclassifications or mis-verifications increase and maintaining similar accuracy on the other legitimate inputs. We try using several different metrics to accomplish these goals (described below).
\end{itemize}

During the search process, each time the network is perturbed, we must test it against a validation set to evaluate if the backdoor is a viable candidate. The $test()$ function in Alg. 1 is where this happens. The process of testing a network is as follows:

The function passes images through the network model and based on the actual class and predicted class of the images, increments different values related to recognition performance. Each probe image is either the attacker, an enrolled entity, or an unknown entity (\textit{i.e.}, some other impostor). If the probe image is from the attacker, then the model is told to verify it as the target. In this case, a decision of true is correct and a decision of false is incorrect. If the probe image is of a legitimate enrolled entity, then the model is asked to verify it against properly matching data. In this case, a decision of true is correct and a decision of false is incorrect. If the image is of an unknown entity, then the model is asked to verify the image as belonging to one of the legitimate enrolled entities. In this case, a decision of true is actually incorrect, and a decision of false is correct. 

These pieces of data determine the model's score based on one of the following metrics in which \textit{wrong} is the total number of incorrect predictions in all classes, \textit{total} is the total number of probe images provided to the model, \textit{I} is the set of all imposter images controlled by the attacker, \textit{K} is the set of all images of known entities, and \textit{U} is the set of all images of unknown entities (\textit{i.e.}, other impostors). For each of the following metrics, a lower score is better.

\begin{equation}
\label{eq:all}
ACC_{all} = \frac{wrong}{total}
\end{equation}

Eq.~\ref{eq:all} is the accuracy over all classes.

\begin{equation}
\label{eq:double}
ACC_{2 \times I_{false}} = \frac{wrong + I_{false}}{total}
\end{equation}

Eq.~\ref{eq:double} more strongly penalizes attacker-related errors.

\begin{equation}
\label{eq:I}
ACC_{all + I} = \frac{wrong}{total} + \frac{I_{false}}{I_{total}}
\end{equation}

Eq.~\ref{eq:I} takes the accuracy over all classes, combined with the accuracy for just the attacker's class.

\begin{equation}
\label{eq:combo}
ACC_{combo} = \frac{I_{false}}{I_{total}} + \frac{K_{false}}{K_{total}} + \frac{U_{true}}{U_{total}}
\end{equation}

Eq.~\ref{eq:combo} is the accuracy of the three categories combined. 

\begin{figure*}[t]
\begin{center}
\includegraphics[width=17cm]{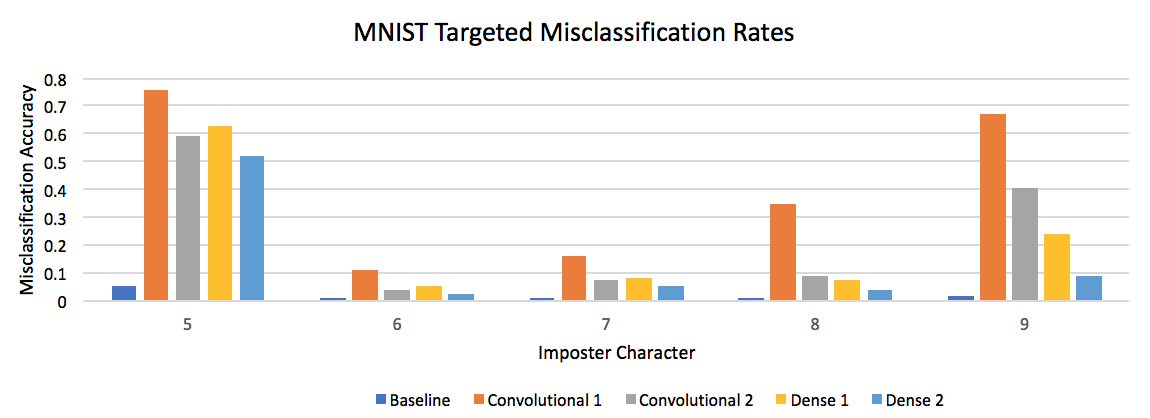}
\end{center}
   \caption{Each bar represents the rate at which a given model misclassifies an attacker specified ``impostor character" as one of the valid characters (the misclassification accuracy). The models are separated by which character is chosen and by which layer of the network was perturbed. Baseline is the rate of misclassification for a given character before any perturbations. The results show that this attack isn't specific to one imposter or one layer, but can be generalized to a variety of choices. All perturbed models maintain accuracy on all other inputs within 0.5\% of the accuracy of the original unaltered network.}
\label{fig:mnist}
\vspace{-3mm}
\end{figure*}

\section{MNIST Digit Classification Attack}
Training and testing CNNs is a computationally expensive task, so as a proof-of-concept before diving into the full task of backdooring a face verification system, we chose to examine a ``toy" scenario incorporating an MNIST digit recognition model, which classifies images of handwritten digits into ten classes representing the numbers 0-9. 

\subsection{MNIST Dataset and CNN Specifics}
For this attack, we used a modified version of an off-the-shelf MNIST classifier. We wanted to mimic our eventual task of an attacker perturbing a facial recognition network and gaining unauthorized access. To do this, we started with a model following the standard architecture for this task, which was obtained from the Keras deep learning library~\cite{kerasModel}. It has two convolutional layers and two fully connected layers. We altered the last layer (the classifier) to output six classes instead of ten. We then retrained the network labeling the digits 0-4 as usual to represent our valid users, and the digits 5-9 as an ``other" category to represent invalid users. A grayscale image of a digit is the input to the model, and the output is the label of the predicted digit as well as the confidence of the model. 

The output of the model is a six element array with the probabilities for each of the known classes as well as the ``other" class. If the highest probability belongs to one of the known classes, it will be accepted and classified as such. If the ``other" category had the highest probability, the image will be rejected. This way of training a model gives it less information than is typically given to MNIST classifiers. This is reflected in the lower level of accuracy achieved on the MNIST test set: 87.9\%. The reason we chose to restrict the information given to the model is that this scenario better simulates the face recognition scenario. The system would not have knowledge of any faces that it has not been trained to classify. 

\subsection{Attack Results}
We show some level of a successful attack for every combination of layer and imposter character in Fig.~\ref{fig:mnist}. To accomplish this, we experimented with the type of perturbation to use as well as how large of a subset of the weights in a layer to perturb. Random and multiplicative perturbations were unsuccessful, so almost all of our attacks used additive perturbations. In general, perturbing between 1\% and 5\% of a given layer's weights was much more successful than targeting a greater portion of the weights. We used Eq. 2 for our metric, because we were able to produce good results without refining the metric at all. Though not flawless, the results of this attack show that our proposed backdoor is a real vulnerability. We were able to produce models that reliably misclassified our imposter character for almost every combination of character and layer perturbed without significantly impacting the accuracy on other inputs. The way we compiled our results was by taking the iteration of the attack that performed the best for a combination of imposter character and layer and graphing it. We define the \textit{best} iteration to be the one that has the highest accuracy of misclassification of the imposter while maintaining accuracy on all other inputs within 0.5\% of the accuracy of the original unaltered network. 

\begin{figure*}[t]
\begin{center}
\includegraphics[width=17cm]{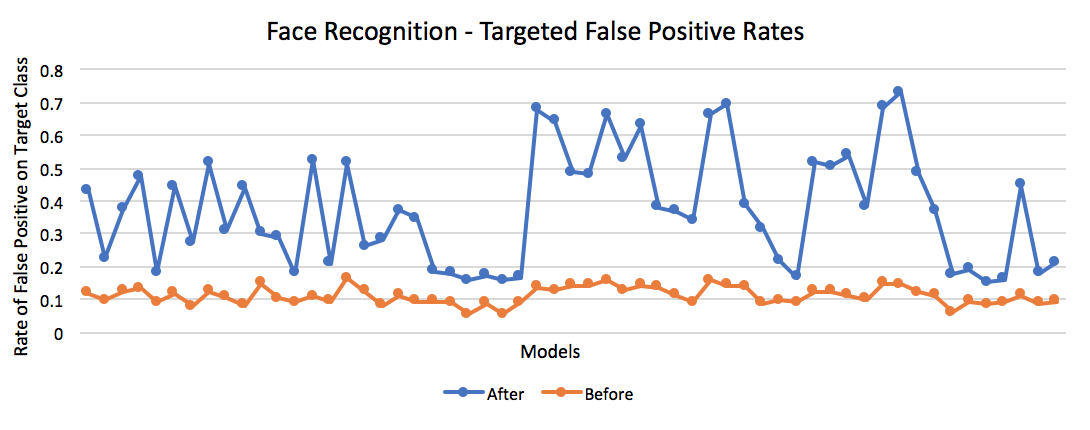}
\end{center}
\vspace{-5mm}
   \caption{A selection of the best backdoor candidates discovered. Each blue point represents a different model perturbed with a different imposter-target pair as its intended false positive scenario. Each orange point represents the original false positive rate for the same imposter-target pair of the original unperturbed model above it. This data shows that this attack generalizes well, succeeding for various identities. Models shown are limited to perturbations that resulted in targeted false positives at a rate of 15\% or greater.}
\label{fig:points}
\vspace{-3mm}
\end{figure*}

 For each experiment we ran approximately 1000 iterations to try to get the best results. Because each time we perturb the network we need to test it over several thousand images, the experiments take several hours to run. Even after this span of time, many of the experiments failed to produce significant results, forcing a change of the hyperparameters and a rerun of the experiment. So an attacker would have to have a substantial amount of time for trial and error to find a model that yields a functional backdoor. However, this is fairly typical of malicious attacks, where patience leads to effective results in more traditional security settings (\textit{e.g.}, password cracking).

We believe that regardless of the imposter character or the layer being perturbed it is possible to make this attack work, but that certain combinations will require more fine-tuning of the parameters of the perturbation algorithm. For example, when perturbing the first convolutional layer, we were able to get high rates of misclassification, but with the second dense layer we were less successful. The same can be seen when looking at the different imposter characters. 

\section{Face Recognition Attack}
Next we turn to a real attack. For face recognition, an attacker perturbs a network with the objective of verifying their face as belonging to someone else without detection.

\subsection{Face Dataset and CNN Specifics}
To witness the full effect of this vulnerability in CNNs, we chose to attack a ResNet50~\cite{he2016deep} architecture trained on the VGGFace2~\cite{cao2018vggface2} dataset. This architecture has 50 convolutional layers that are organized into 16 blocks. We made use of an implementation for the Keras deep learning library~\cite{kerasModel2} that had the option to use weights pre-trained on the VGGFace2 dataset. We chose to use these pre-trained weights, but used a variation of the network architecture that excluded the fully connected layers associated with classification at the top of the network. This version of the network takes a 224x224 RGB image as input and outputs a 2048-dimensional feature vector. So instead of making a prediction in the set of known faces, the network outputs a feature-vector  for the image that has been presented to it. 

To setup the verification system, we downloaded approximately 160,000 images of 500 distinct subjects from the VGGFace2 dataset. We input the first 100 images of each subject to the network and averaged together the outputs for each to give us our enrollment of that subject. To verify an input image as the claimed identity, we use cosine similarity to compare the image's feature vector to the average vector that we have stored for that subject. We iterated over 100 more images for each subject, and compared each image to the stored vector for each subject. We then chose a threshold for cosine similarity that would attempt to maximize true positives and minimize false positives. When a user claims a certain identity for an image, if the cosine similarity between that image and the representation of that subject is above the threshold, the model outputs true, else it outputs false. This network and threshold achieved an accuracy of 88.7\% on the VGGFace2 test set.

\subsection{Attack Results}
 The network is much deeper than in the MNIST case study, and the input images are larger and in RGB colorspace instead of grayscale. Since each iteration has to be tested on approximately 10,000 images, running a couple hundred iterations of one of these experiments takes between 12 and 18 hours --- even when running the tests on GPU systems. For this reason we chose to drastically reduce our search space.
 
 We wanted to show that many different imposters could be verified as many different targets, so we decided to perturb the same layer for all of our experiments. Once we had our layer, we used different subset sizes and objective metrics until we found a combination of hyperparameters that seemed to work for many different pairs of attacker controlled imposters and targets. This reduced the dimensionality of the experiment by quite a lot. We chose to use the first convolutional layer of the network for this attack since the first layer of the MNIST experiment was by far the most successful. We perturbed 1\% of the weights in this layer each time, because larger fractions seemed to alter the behavior too significantly and smaller fractions did not seem to have a significant enough effect. Lastly, after a few runs with different objective functions, we chose to use Eq. 3 from Sec. 3 for the objective function. It appeared to put the correct amount of weight on each component of the model's accuracy. Using Eq. 4 as the metric also seemed to produce good results in some cases, but we wanted to limit the number of variations between each test.

\begin{figure}[t]
\begin{center}
\includegraphics[width=8cm]{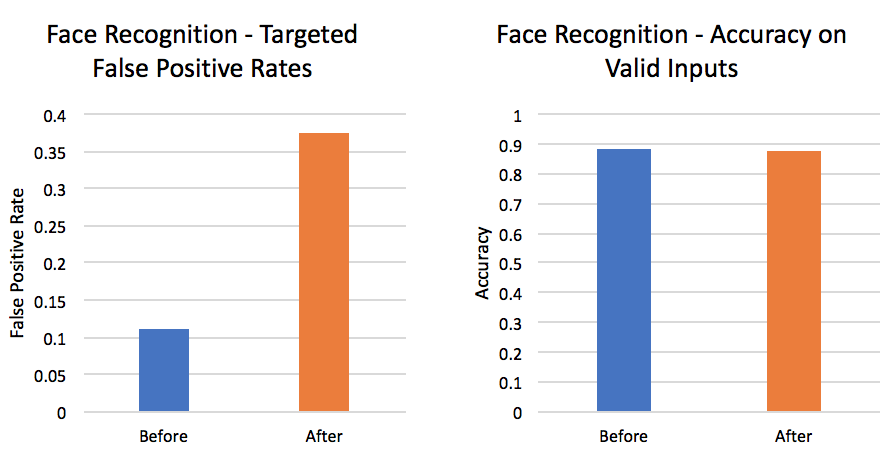}
\end{center}
   \vspace{-3mm}
   \caption{Average performance over all models before and after perturbations. The left graph is the false positive rate for the intended imposter-target pairs. The right graph is the accuracy on all other inputs. Our algorithm was successful in supporting targeted false positives while maintaining a high level of performance for non-target inputs. Models were limited to perturbations that resulted in targeted false positives at a rate of 15\% or greater.}
\label{fig:avg}
\vspace{-4mm}
\end{figure}

This setup was used to test 150 different imposter and target pairs, running 300-400 iterations for each pair. Even in this limited setup, 38\% of the pairs yielded plausible results, and 15\% of the pairs yielded successful results. We define plausible as a model that outputs false positives for our specific target class greater than 15\% of the time while keeping accuracy of the model on all other inputs within 1.5\% of its original accuracy. A successful attack is the same, but with a false positive rate of at least 40\% on the target class. A few models even showed mis-verification rates as high as 75\%. Fig.~\ref{fig:points} shows the false positive rate for a target class before and after perturbing a network for all plausible and successful iterations. All points on the \textit{before} line are representative of the original network, whereas points on the \textit{after} line each represent a different perturbed version of the model. Fig.~\ref{fig:avg} shows the average performance over all of these models. Based on these findings, we are confident that we can always find a combination of hyperparameters that leads to a high success rate for an arbitrary impostor and target pair. 

\section{Discussion}
A very useful property of artificial neural networks is that they can learn the same arbitrary function in more than one way. This flexibility allows us to change training data and hyperparameters at will, but still learn a set of weights that solves a specific problem like face verification for a set of known identities at a target accuracy rate. However, in a security context, this property can be a liability. Our results lead us to believe that with the right amount of time and knowledge of the network, an attacker can successfully alter the weights of a pre-trained CNN model in such a way as to be able to reliably mis-verify an imposter class as some other legitimate class. 

But surely there are some viable defenses against this backdoor attack. A straightforward way to detect an attack like this one is to periodically compute a one-way hash function against the model's underlying file on the system and check it against a known good hash for that file. If the computed hash and the stored hash are not the same, then we know the model has changed in some, possibly malicious, way. Of course, this strategy is not foolproof, and may not be reliable is several different scenarios. 

First, networks with stochastic outputs are becoming more common as machine learning practitioners seek to understand the reliability of their models. There are two ways this is commonly implemented: small random weight perturbations at test time~\cite{NIPS2011_4329,goodfellow2016deep} and dropout at test time~\cite{pmlr-v48-gal16}. Both can change the stored representation of the network on disk, thus rendering the hash verification completely ineffective (the hashes will always be mismatched). 

Second, depending on the access the attacker has to the operating system the CNN is running on top of, it is possible to turn to traditional rootkits that manipulate system calls to misdirect the one-way hash function to a preserved original network file when an attempt is made to validate the backdoored network file. This is a classic, yet still useful, trick to defeat such host-based intrusion detection strategies.

Finally, if a weak hash function (\textit{e.g.}, MD5, SHA1) has been chosen it is conceivable that an extra constraint could be added to the backdoor search process that looks for weight perturbations that not only maximize the attacker's chances of authenticating and minimize the impact on legitimate users, but also yield a useful hash collision. This means the backdoored network file would produce the same exact hash as the original network file.
 
Thus we leave the reliable detection and removal of backdoors from targeted weight perturbations as an open question. In principle, there should be a way to better characterize failure modes of CNNs beyond dataset-driven error rates. Recent work in explainable AI might be one direction that is helpful. Further, network fine-pruning~\cite{DBLP:journals/corr/abs-1805-12185} has been demonstrated to be an effective mechanism to remove other forms of backdoors in CNNs. Such a strategy might also help in the case of the attack we propose here, although the attacker can always respond with a new set of perturbed weights after fine-pruning.  We hope that this study may add to the increasing understanding of the security of artificial neural networks. To encourage further work, code and data associated with this paper will be released upon publication. 

{\small
\bibliographystyle{ieee}
\bibliography{references}
}

\end{document}